\documentclass[10pt, a4paper]{article}

\usepackage[final]{lrec2026} 
\usepackage{subcaption}
\usepackage{amsmath}
\usepackage{booktabs}
\usepackage{hyperref}

\title{The Patrologia Graeca Corpus: OCR, Annotation, and Open Release of Noisy Nineteenth-Century Polytonic Greek Editions}

\name{Chahan Vidal-Gorène$^{1,2}$, Bastien Kindt$^{3}$} 

\address{$^{1}$École nationale des chartes -- PSL, Paris, France \\$^{2}$Calfa, Paris, France \\ chahan.vidal-gorene@chartes.psl.eu
          \\\\
         $^{3}$UCLouvain -- CIOL Institut orientaliste, Louvain-la-Neuve, Belgium \\ bastien.kindt@uclouvain.be \\}

\abstract{
We present the \textit{Patrologia Graeca Corpus}, the first large-scale open OCR and linguistic resource for nineteenth-century editions of Ancient Greek. 
The collection covers the remaining undigitized volumes of the \textit{Patrologia Graeca} (PG), printed in complex bilingual (Greek–Latin) layouts and characterized by highly degraded polytonic Greek typography. 
Through a dedicated pipeline combining YOLO-based layout detection and CRNN-based text recognition, we achieve a character error rate (CER) of 1.05\% and a word error rate (WER) of 4.69\%, largely outperforming existing OCR systems for polytonic Greek. 
The resulting corpus contains around six million lemmatized and part-of-speech tagged tokens, aligned with full OCR and layout annotations. 
Beyond its philological value, this corpus establishes a new benchmark for OCR on noisy polytonic Greek and provides training material for future models, including LLMs.
 \\ \newline 
}

\begin{document}

\maketitleabstract
\section{Introduction}

Despite major advances in digital philology, Ancient Greek scholarship still lacks a reliable, open, and linguistically enriched corpus covering the \textit{Patrologia Graeca} (PG; 161 vols., 1\textsuperscript{st}–15\textsuperscript{th} c.). Compiled by Jacques-Paul Migne in Paris between 1857 and 1866, the PG reprints a vast range of patristic, exegetical, historical, hagiographical, legal, encyclopedic, poetic, and even narrative texts, encompassing much of Byzantine Greek literature. Yet some of these works have not been re-edited since the nineteenth century and remain unavailable in a machine-readable format. The series is widely accessible as PDF scans (e.g., patristica.net, archive.org, books.google.com portals or Roger Pearse's blog) and lacks structural encoding, which severely limits textual reuse, searchability, and linguistic analysis~\citeplanguageresource{patristica,pearse-pg-pdfs}.

The absence of a structured, open, and linguistically annotated corpus of the PG prevents the application of computational approaches to a significant portion of Greek literature, as well as the development of models representing the full diachronic and stylistic range of the language. While the \citetlanguageresource{scaife-about} (Perseus 5; part of the Open Greek and Latin Project, with initial development led by Eldarion and funded by the Alexander von Humboldt Chair of Digital Humanities at Leipzig), the \citetlanguageresource{pta-site}, and the \citetlanguageresource{tlg-abridged} have transformed access to ancient texts, they remain partial or restricted for the post-classical and Byzantine periods. 

This project addresses these gaps by identifying, digitizing, and linguistically annotating the still-unavailable volumes of the PG. The resulting corpus—around 6,000,000 lemmatized and morphologically parsed words—substantially extends the lexical and stylistic range of available Ancient Greek resources. Thanks to the wide thematic scope and chronological span of the PG, the dataset contributes thousands of rare inflected forms, technical and theological terms, named entities, and toponyms that are underrepresented in existing corpora, thereby providing a stronger coverage for Ancient Greek language models in the future.

\section{Previous Initiatives and the Challenges of the \textit{Patrologia Graeca}}

\subsection{Digital initiatives around the PG}

Previous attempts to digitize the \textit{Patrologia Graeca} (PG) have primarily consisted in providing PDF scans without reliable text extraction or structural markup, notably through \citetlanguageresource{patristica} and related initiatives curated by Pearse. Beyond these, only partial textual coverage exists in scholarly databases such as the \citetlanguageresource{tlg-abridged} and in open-access repositories like the \citetlanguageresource{pta-site} (composed mainly of texts from the \textit{Griechischen Christlichen Schriftsteller} series).

The most ambitious effort to date was the \textit{Open Patrologia Graeca 1.0} (OPG) project~\cite{perseus-opg}. Launched in 2015 by Gregory Crane and collaborators within the Perseus Digital Library and the Open Greek and Latin consortium, it aimed to produce OCR-derived Greek and Latin texts for all 161 PG volumes—roughly 50 million words—using open-source engines such as OCRopus~\cite{breuel2008ocropus} and Tesseract. The raw outputs, released openly on GitHub under \citetlanguageresource{oglpg-github}, represented an unprecedented attempt at large-scale text recovery for late antique and Byzantine Greek. While the project demonstrated the feasibility of large-scale OCR for polytonic Greek, the output quality remained highly variable, with frequent segmentation errors (distinction of latin lines vs greek lines, but also segmentation or characters), inconsistent encoding, and minimal metadata. The data were never converted into TEI-structured or linguistically annotated form, and the repositories have remained unmaintained since 2017. Later analyses confirmed that the OCR accuracy on the PG corpus is typically below 90\% and that recognition performance degrades sharply for small fonts and complex diacritics~\cite{robertson2017large,varthis2022semantic}. These limitations illustrate both the promise and the fragility of fully automatic approaches to nineteenth-century Greek printing.

In summary, the digital landscape of the \textit{Patrologia Graeca} remains fragmented: most volumes exist only as non-searchable PDF scans; large-scale OCR attempts have produced uncorrected and inconsistent text; and a few TEI-encoded editions cover isolated works.

\subsection{OCR and Editorial Challenges}

\begin{figure}[!hb]
\begin{center}
\includegraphics[width=\columnwidth]{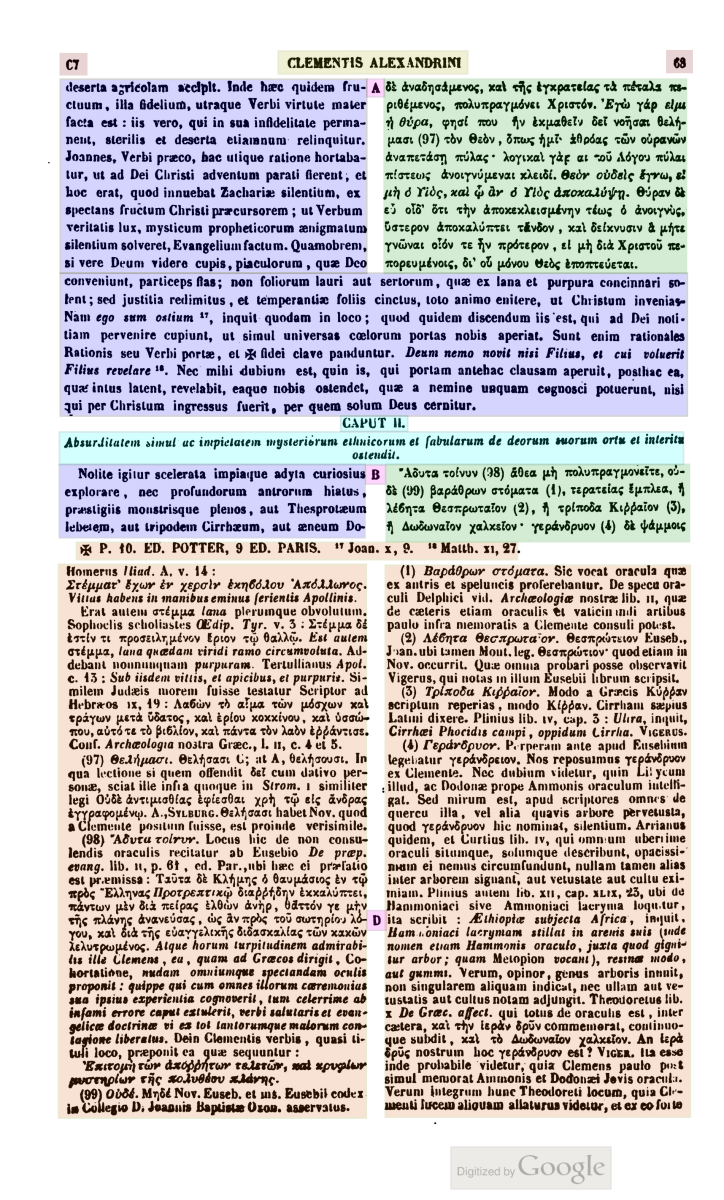}
\caption{Example of page layout in PG with different semantic zones. Green regions correspond to the targeted Greek text.}
\label{fig:layoutPG}
\end{center}
\end{figure}

The \textit{Patrologia Graeca} presents exceptional challenges for both OCR and subsequent linguistic analysis. Each volume reproduces heterogeneous nineteenth-century reprints, typically (but not constantly) arranged in two columns (Greek and Latin, often with overlapping content flowing from one language to the other), and surrounded by marginalia, running titles, and apparatus notes (see Fig.~\ref{fig:layoutPG}).

\begin{figure}[!ht]
\begin{center}
\includegraphics[width=\columnwidth]{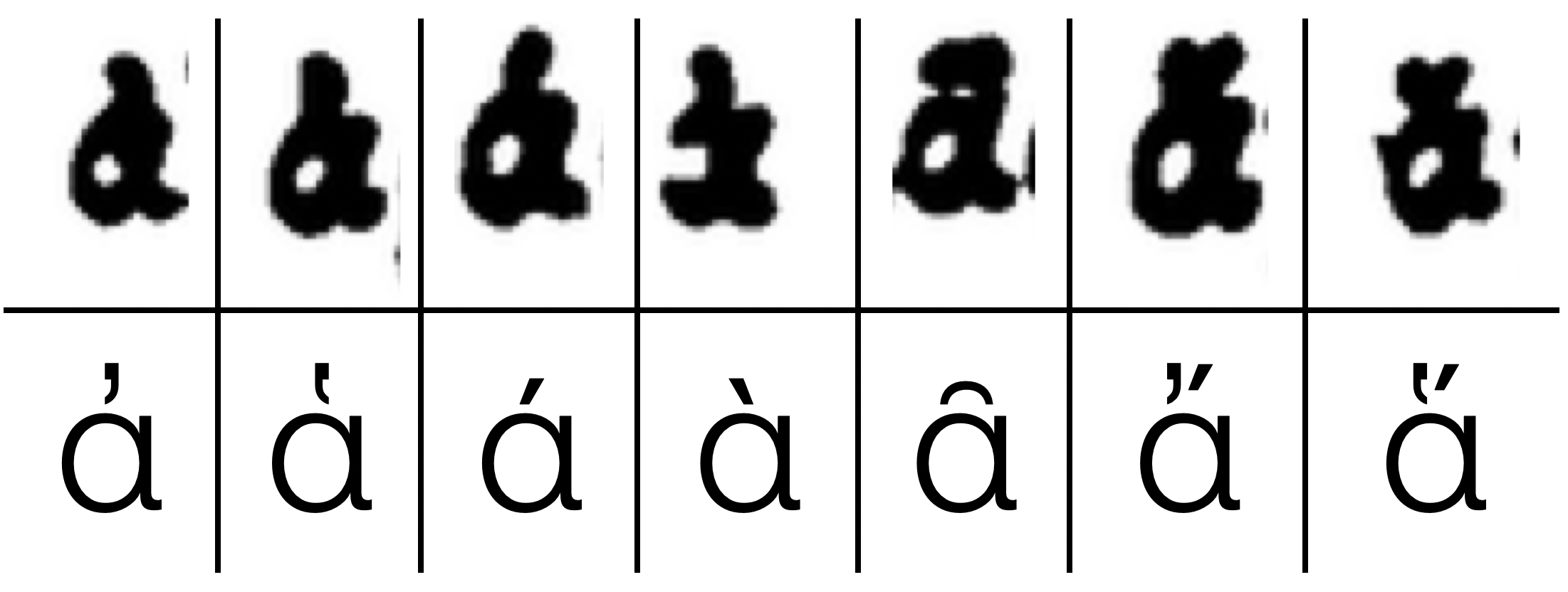}
\caption{Typography ambiguity in PG, variation of the character \textit{alpha} with diacritics. See Section~\ref{sec:results_and_corpus_release} for impact on OCR results}
\label{fig:TypoPG}
\end{center}
\end{figure}

Typographical quality varies greatly from one volume to another, and the dense use of polytonic diacritics severely complicates line segmentation and character recognition~\cite{varthis2022semantic,vidal2023reconnaissance}. Common scanning issues—page curvature, misalignment, and excessive binarization—further degrade image quality, increasing ambiguity in reading (see Fig.~\ref{fig:TypoPG}). Moreover, no standardized, page-aligned ground truth datasets exist for these materials, which prevents the training and evaluation of dedicated OCR or layout analysis models for complex polytonic Greek~\citep{robertson2017large,romanello-2021}.

\section{Related Work in Greek OCR and Document Analysis}

Research on Greek OCR has primarily focused on the structural complexity of scholarly editions. Early work on the \textit{Patrologia Graeca}~\cite{robertson2014automated} addressed column segmentation through geometric preprocessing—detecting intercolumn regions via Hough transforms and filtering spurious letters with a k-nearest-neighbor classifier—thus stabilizing OCR input. Later studies on dense critical layouts~\cite{sven2022page} confirmed that purely visual object-detection models (notably YOLO) outperform textual or hybrid approaches (LayoutLMv3, RoBERTa), since segmentation in such editions depends more on spatial than linguistic features.

Tesseract, trained mainly on synthetic data, performs poorly on the \textit{Patrologia Graeca}~\cite{perseus-opg,varthis2022semantic}. Alternative approaches such as \textit{word spotting}~\cite{varthis2022semantic} work well on specific cases but lack generalization. More recently, page-level vision–language models\cite{semnani2025churro} have introduced contextual and multilingual OCR, capable of bypassing explicit segmentation.

Fine-tuning such models, however, requires extensive data, while existing corpora for polytonic Greek offer mostly line-level ground truth: \texttt{greek\_cursive} from \citetlanguageresource{ocr-greekcursive}; \texttt{gaza-iliad}, \texttt{voulgaris-aeneid}, and \texttt{gaza-batrachomyomachia} from \citetlanguageresource{baumann2022a,baumann2022b,baumann2023}; \texttt{GT4HistComment}~\cite{romanello-2021}; \texttt{Pogretra} from \citetlanguageresource{bruce-robertson-2021-4774201}; and \texttt{GRPOLY-DB}~\cite{7333841}. These datasets provide strong baselines but remain narrowly focused (clean fonts, multilingual data; see \citet{vidal2023reconnaissance}). Though unsuited to the thick and ambiguous typography of the \textit{Patrologia Graeca}, GT4HistComment and Pogretra replicate more realistic scholarly layouts and supply useful lexical diversity for foundation model pretraining.

Automatic text recognition (OCR/HTR) is now standard in large-scale digitization, notably through \textit{Transkribus}~\cite{kahle2017transkribus}, \textit{eScriptorium}, and \textit{Calfa Vision}~\cite{vidal2021modular}. These platforms democratize model training by providing integrated fine-tuning pipelines based on baseline-oriented layout annotation and deep neural architectures, including fine-tunable CNN–LSTM and transformer-based networks~\cite{reul2019ocr4all,vidal2021modular}. In particular, active-learning and automated iterative fine-tuning strategies implemented in Calfa Vision have yielded major gains on Greek, reducing CER by 63.94 points after only ten pages of transcription~\cite{vidal2023reconnaissance}. The platform’s modular design also allows integration of YOLO-based layout models, identified as the most effective for these tasks (layout analysis and line detection, without baseline)~\cite{sven2022page}.

\smallskip
For lemmatization, POS tagging, and morphological parsing, we adopt the hybrid strategy developed for the Byzantine corpus \textit{De Thessalonica Capta}~\cite{kindt2022analyse}, combining neural tagging and rule- or dictionary-based post-correction. This method, built upon a fine-tuned version of the PIE architecture~\cite{manjavacas2019improving} adapted to the GRE\textit{g}ORI tagset~\citeplanguageresource{clerice-pie}, offers higher robustness for highly inflected and diachronic Greek. It reaches 97.94\% lemmatization and 98.59\% POS-tagging accuracy, while maintaining interpretability and stable handling of polylexical and enclitic forms.

\smallskip
This hybrid workflow also serves a corpus-building objective: it will yield approximately six million additional words of linguistically normalized Greek, thereby extending existing resources for model training. Such enrichment directly benefits the next generation of Ancient-Greek transformer models—e.g. \textit{Ancient-Greek-BERT}~\cite{ancient-greek-bert}, \textit{GreBERTa} and \textit{PhilBERTa}~\cite{riemenschneiderfrank:2023}, or \textit{LOGION}~\cite{logion-base}—whose performance depends on lexically diverse, diachronic data. Our corpus thus plays a dual role: improving automatic annotation quality through a dedicated tagset, and contributing new training material for future re-training or benchmarking of Ancient Greek BERT-like architectures.

\smallskip
Beyond OCR, recent work has explored the semantic enrichment of Greek and Latin corpora through translation alignment, syntactic annotation, and named entity recognition. Projects such as Ugarit and \textit{Beyond Translation}~\cite{Crane2025} combine manual and automatic alignments with treebank data, enabling large-scale semantic and cross-lingual analysis. The creation of large, linguistically normalized corpora—such as the extended PG dataset proposed here—provides the substrate required for robust adaptation of these models to post-classical and Byzantine Greek.

\section{Methodology and dataset for PG OCR and analysis}
\subsection{Data Preparation}
To evaluate OCR performance on the \textit{Patrologia Graeca} (PG), we first established a 30 pages test set, randomly sampled from the corpus and manually transcribed for the Greek text regions. Existing public models for Greek---namely the \textit{tesseract-ocr}\footnote{\url{https://github.com/tesseract-ocr/tessdata_best/blob/main/grc.traineddata}} and \textit{Transkribus} model for nineteenth-century Greek prints\footnote{\url{https://app.transkribus.org/models/public/text/19th-century-greek-8.0}}---fail to achieve an average character error rate (CER) below 5\% (see Table~\ref{tab:ocr-results}). This confirms that PG presents challenges beyond the scope of existing models, due to its complex typography, heterogeneous layouts, and degraded scan quality.

\begin{figure*}[!ht]
\begin{center}
\includegraphics[width=\textwidth]{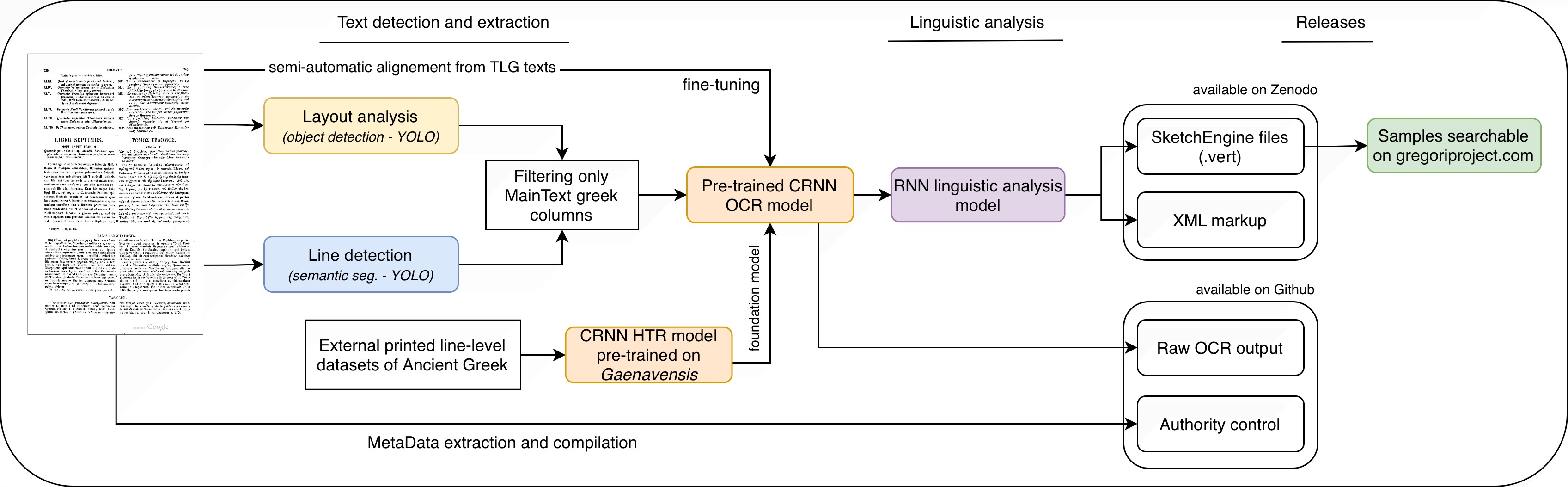}
\caption{Overview of the OCR and annotation workflow for the \textit{Patrologia Graeca} corpus.}
\label{fig:pipelinePG}
\end{center}
\end{figure*}

\subsection{Model Architecture and Fine-tuning}

\smallskip
To address this, we used a pretrained word-based CRNN architecture on a Greek manuscript, the \textit{Genavensis Græcus 44}. This model was then fine-tuned on printed datasets previously described, augmented with 50k synthetic samples generated through the \textit{albumentations} library in order to create a foundation model for our task. We applied aggressive degradations (Gaussian noise, motion blur, compression artifacts, contrast shifts, elastic distortion, random fog, rain, and coarse dropout) to reproduce the visual variability and damage patterns typical of PG scans. While this foundation model did not yet reduce the CER significantly, it stabilized the word error rate (WER) around 10\% thanks to improved lexical coverage, and served as a robust starting point for PG-specific fine-tuning (see Table~\ref{tab:ocr-results}).

\smallskip
The main PG training dataset was then created following the iterative fine-tuning strategy of \citet{vidal2023reconnaissance}. Segments of the PG already available in digital form through the \citetlanguageresource{tlg-abridged} were semi-automatically aligned within the Calfa Vision platform~\cite{vidal2021modular}, combining automatic layout detection with manual correction. The resulting dataset comprises 445 images representative of the typographical and structural diversity of the PG (see Table~\ref{tab:groundtruth}). A specific focus has been made on pages with crossing text-regions (see Figure~\ref{fig:layoutPG} with the Latin column in blue).

For text analysis, we use a model trained on a representative sample of the corpora from the GRE\textit{g}ORI database~\cite{Kindt2025-GREGORI}, which comprises more than two million lemmatized word forms covering Classical and Byzantine Greek from the first to the fifteenth century, lemmatized and morphosyntactically annotated through rule-based dictionary methods and manually corrected. The model used is the one trained by \citet{kindt2022analyse} for the \textit{De Thessalonica Capta} corpus. Figure \ref{fig:pipelinePG} illustrates the processing pipeline.

\begin{table}[!hb]
\centering
\small
\begin{tabular}{l r}
\hline
\textbf{Global information} & \textbf{Count} \\
\hline
Images & 445 \\
TextLines & 11,096 \\
LinePolygons & 6,921 \\
TextRegions & 4,342 \\\\
\textbf{Semantic categories} & \textbf{Count} \\
\hline
MainText\_ColGreek & 1,022 \\
MainText\_ColLatin & 761 \\
MainText\_Title & 267 \\
Marginalia & 476 \\
Marginalia\_Footnote & 652 \\
Marginalia\_PageNumber & 611 \\
Marginalia\_ParagraphNumber & 248 \\
Title\_RunningTitle & 305 \\
\hline
\end{tabular}
\caption{OCR ground-truth}
\label{tab:groundtruth}
\end{table}

\section{Results and Corpus Release}\label{sec:results_and_corpus_release}
\subsection{OCR and Layout Performance}
The results in Table~\ref{tab:ocr-results} summarize the performances in text recognition and layout analysis. Our OCR model, specifically fine-tuned for the PG material, clearly outperforms all previously used systems on this corpus, as well as a model trained on synthetic data, achieving a gain of 5–7 points in CER and 6–10 points in WER.

\begin{figure*}[!ht]
\begin{center}
\includegraphics[width=\textwidth]{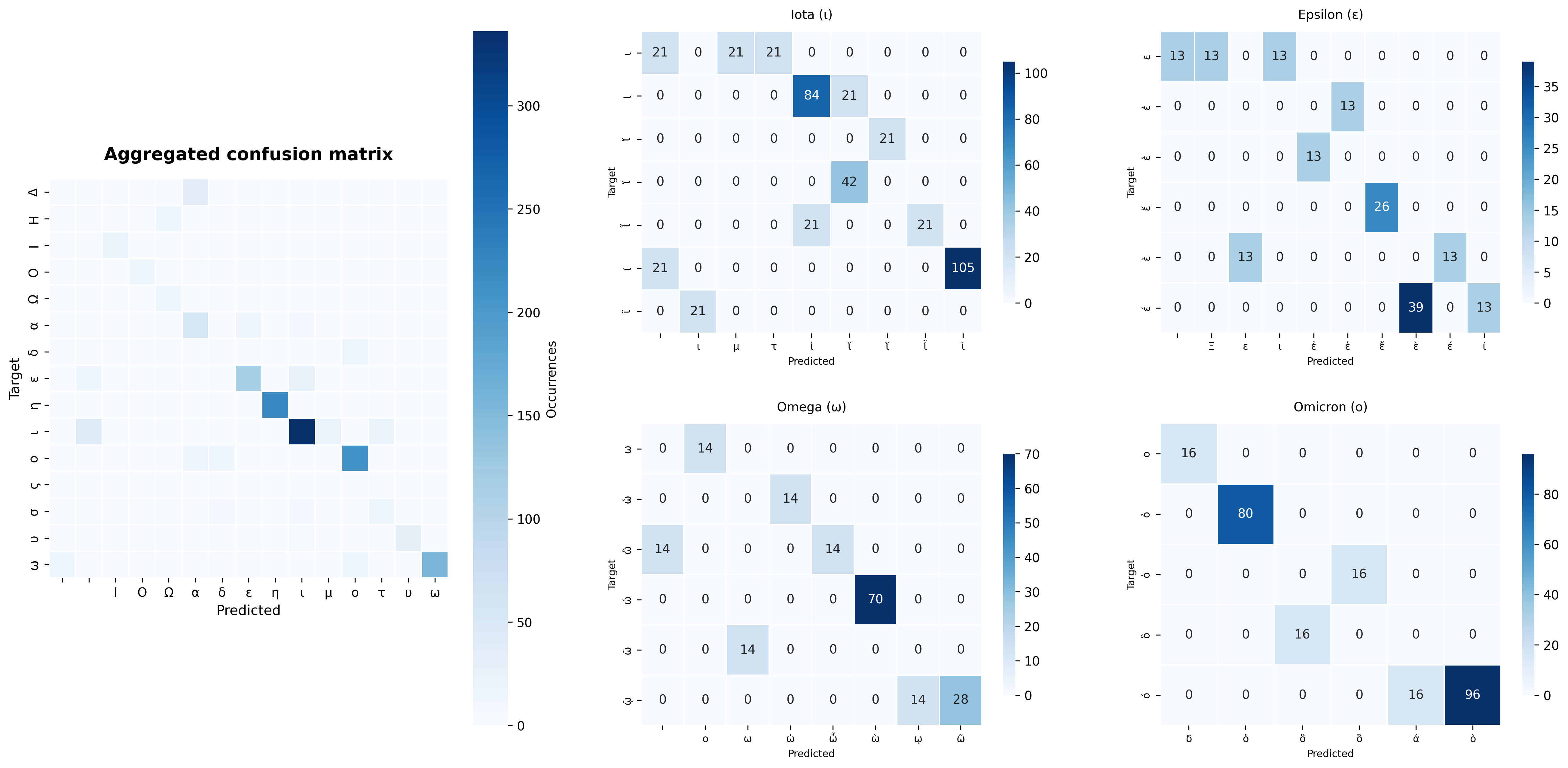}
\caption{Character-level confusion matrix (left) and detailed confusion patterns for iota, omega, epsilon, and omicron showing diacritic variation}
\label{fig:confusion}
\end{center}
\end{figure*}

The most frequent OCR errors concern diacritic confusions among polytonic Greek characters.
The majority of mismatches [e.g., $\acute{\iota}$ (03Af) / $\acute{\iota}$ (1F77), $\acute{\alpha}$ (03AC) /$\acute{\alpha}$ (1F71), $\acute{\epsilon}$ (03AD) / $\acute{\epsilon}$ (1F73), 
$\acute{o}$ (03CC) / $\acute{o}$ (1F79), $\acute{\upsilon}$  (03CD) / $\acute{\upsilon}$ (1F7B), $\acute{\eta}$ (03AE) / $\acute{\eta}$ (1F75), $\acute{\omega}$ (03CE) /$\acute{\omega}$ (1F7D)] correspond to visually identical characters whose accent encodings differ, typically between monotonic and polytonic forms. This is due to non-homogenity in the datasets used. These substitutions account for over 80\% of all errors, showing that the model usually recognizes the correct letter identity but fails to reproduce the exact diacritic.

A closer examination of the 123 distinct error patterns observed across 1,572 total occurrences reveals systematic challenges by letter group (Fig.~\ref{fig:confusion}). \textit{Iota} exhibits the highest error rate (315 occurrences), primarily due to confusion between rough and smooth breathing marks (e.g., different diacritics on $\iota$), followed by omission of diacritics entirely. Interestingly, iota is also occasionally misread as morphologically unrelated characters such as $\tau$ or $\mu$, suggesting low-level visual similarity in poorly inked regions. \textit{Omicron} (144 occurrences) shows predominantly breathing-related errors, with a few instances of cross-letter confusion with $\delta$ or misrecognition of uppercase forms. \textit{Omega} (126 occurrences) presents similar diacritic instability, especially in subscript iota and accent type, with one notable cross-vowel error reflecting visual proximity in compressed typography. \textit{Epsilon} (117 occurrences) exhibits the most diverse error patterns: breathing confusions, total diacritic loss, and a striking confusion with uppercase $\Xi$ likely due to similar vertical strokes in certain typefaces.

A smaller set of errors involves spacing and punctuation, final sigma versus punctuation ($\varsigma \to$ comma or period), inter-consonant confusions ($\sigma/\delta$, $\sigma/\tau$, $\rho/o$), and uppercase mismatches ($\Delta/\Lambda$, $\Psi/T$), accounting for fewer than 10\% of total errors but remaining problematic for lemmatization pipelines.

Overall, most transcription errors can be mitigated through systematic Unicode normalization before training and post-correction rules.
This leads to a substantially lower CER. Errors are still more frequent in uppercase characters (therefore mainly titles), which would require specific reinforcement during training. Post-processing is a necessary task, because if the symbol appears similar, it penalizes text analysis and keyword research. Only Greek columns and titles are relevant in our case; other regions are excluded from the pipeline.

\begin{table}[!ht]
\centering
\small
\resizebox{\columnwidth}{!}{%
\begin{tabular}{lccc}
\multicolumn{4}{c}{\textbf{Text recognition}} \\\\
\hline
\textbf{Model} & \textbf{CER (\%)} & \textbf{WER (\%)} &  \\
\hline
Tesseract (Greek) & 11.57 & 39.65 &  \\
Transkribus (19th~c. Greek) & 6.14 & 14.82 &  \\\\
Pre-trained CRNN \scriptsize{(with artificial noise)} & 8.12 & 11.23 &  \\
\textbf{Ours (PG fine-tuned)} & \textbf{1.05} & \textbf{4.69} &  \\\\

\multicolumn{4}{c}{\textbf{Layout detection}} \\\\
\hline
\textbf{Class} & \textbf{P} & \textbf{R} & \textbf{mAP50} \\
\hline
\textbf{MainText\_ColGreek} & \textbf{0.963} & \textbf{0.994} & \textbf{0.973} \\
MainText\_ColLatin & 0.969 & 1.000 & 0.982 \\
MainText\_Title & 0.462 & 0.950 & 0.462 \\
Marginalia & 0.422 & 0.891 & 0.427 \\
Marginalia\_Footnote & 0.493 & 0.975 & 0.511 \\
Marginalia\_PageNumber & 0.382 & 0.783 & 0.380 \\
Marginalia\_ParagraphNumber & 0.353 & 0.679 & 0.408 \\
Running & 0.477 & 0.883 & 0.542 \\\\
\multicolumn{4}{c}{\textbf{Line detection and reading order}} \\\\
\hline
\textbf{Task} & \textbf{P} & \textbf{R} & \textbf{mAP50} \\
\hline
Line detection & 0.983 & 0.994 & 0.973 \\
Reading order & 0.98 & - & -

\end{tabular}
}
\caption{OCR performance (layout and recognition) comparison on the 30-page PG test set.}
\label{tab:ocr-results}
\end{table}

Beyond the quantitative results on the test set, applying the models in real production conditions reveals additional sources of error. Although it occurs only occasionally, a few cross-column regions in Latin may be mistakenly detected, introducing noticeable noise in the subsequent text recognition. The most persistent challenge, however, concerns title areas, which are not always clearly distinguished between Greek and Latin scripts. Another issue sometimes observed in line detection—particularly in overlapping or crossing regions—is the splitting of a single line into two fragments with their reading order reversed.
To mitigate these residual problems, we adopt an active learning strategy, continuously correcting and reinjecting misrecognized samples into the training data to progressively improve model robustness.

After applying the OCR models to the entire set of PG volumes and processing the recognized text, we obtain a corpus of approximately six million words.

\subsection{Corpus Characteristics and Visualization}

\begin{figure}[!b]
  \centering
  \begin{subfigure}[t]{\columnwidth}
    \includegraphics[width=\columnwidth]{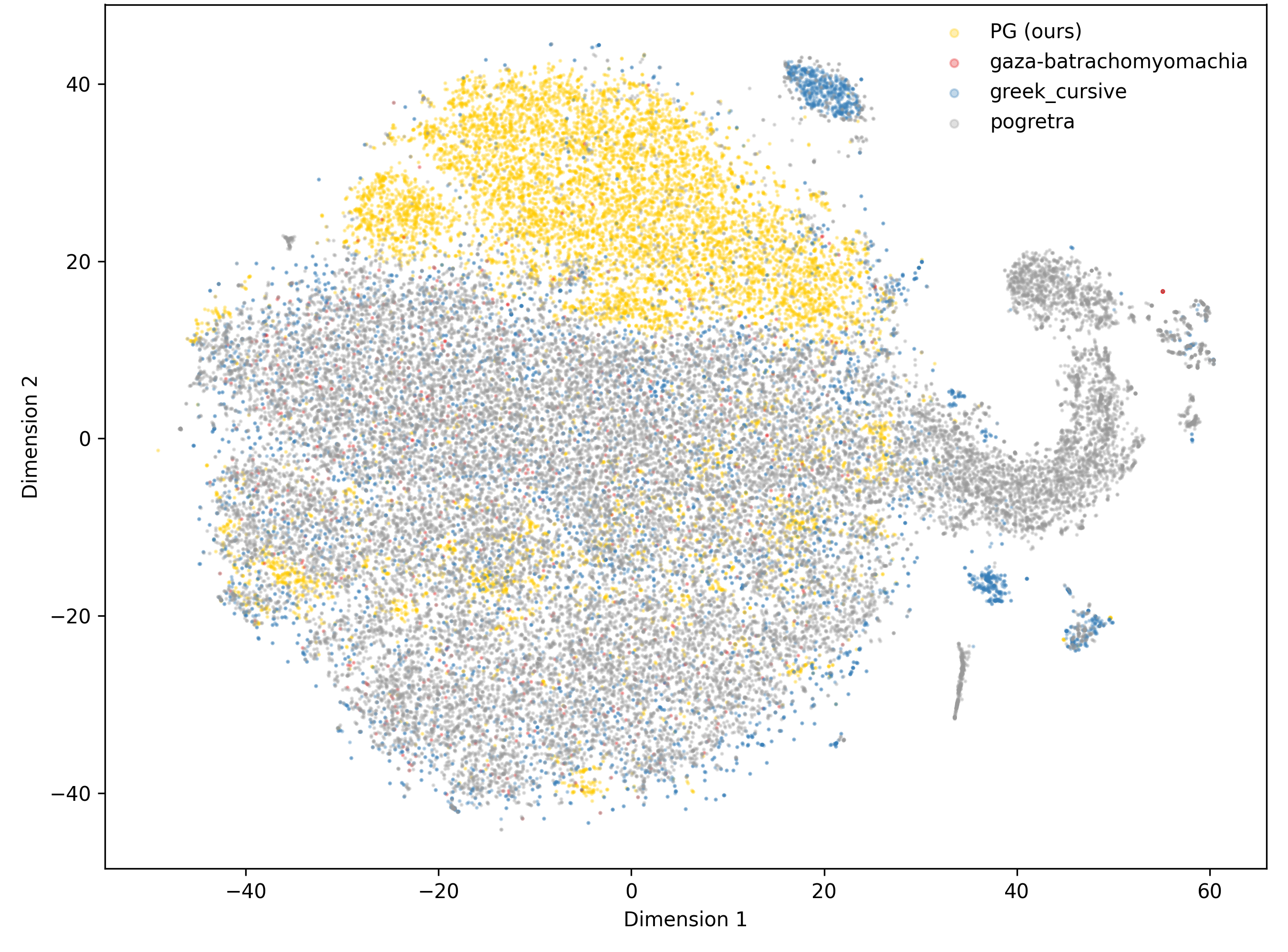}
    \caption{t-SNE of fastText embeddings (semantic)}
  \end{subfigure}\hfill
  \begin{subfigure}[t]{\columnwidth}
    \includegraphics[width=\columnwidth]{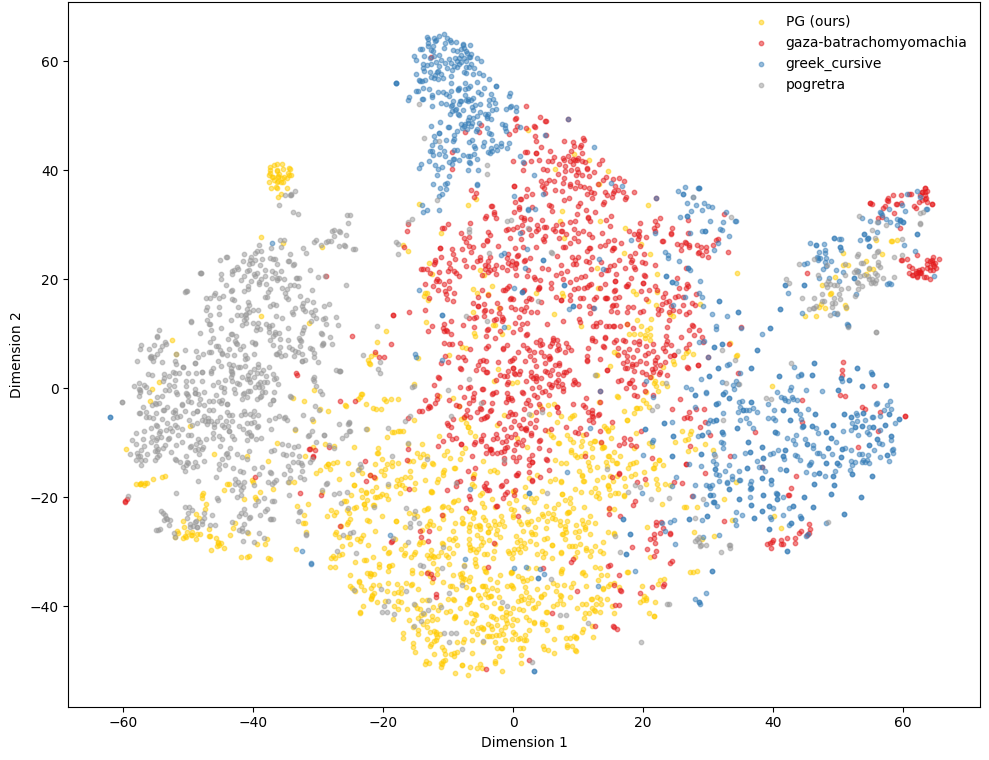}
    \caption{t-SNE on VGG16 embeddings (vision)}
  \end{subfigure}
  \caption{Comparison of semantic and visual distributions of ancient Greek corpora.}
  \label{fig:tsne_comparison}
\end{figure}

\smallskip
To evaluate both the linguistic and visual characteristics of the corpus we release, we projected the existing open corpora into two embedding spaces (Fig.~\ref{fig:tsne_comparison}). Subfigure (a) shows a t-SNE visualization of sentence-level embeddings derived from \textit{fastText} models trained on Ancient Greek. The PG corpus (yellow) forms a coherent but distinct cluster, confirming its lexical specificity compared to other datasets. In contrast, the Pogretra data (grey) appear more dispersed, reflecting OCR noise and inconsistent orthography, while \textit{Gaza-Batrachomyomachia} and \textit{Greek\_cursive} occupy peripheral, stylistically homogeneous regions.

Subfigure (b) presents a t-SNE projection of line-level image embeddings extracted from VGG16 (pretrained CNN). Unlike in the lexical space, the visual distribution shows strong inter-corpus separation, highlighting typographic and palaeographic variation. The PG lines occupy a dense and compact region, visually distinct from both manuscript and modern print corpora. These observations underline the atypical nature of the PG—typographically complex, lexically rich, and divergent from available Greek OCR datasets—thus motivating the creation of a dedicated resource and the development of tailored OCR, normalization workflows and motivating the release of this new corpus for Ancient Greek studies.

\subsection{Output Format and Public Release}
\begin{figure}[!ht]
    \includegraphics[width=\columnwidth]{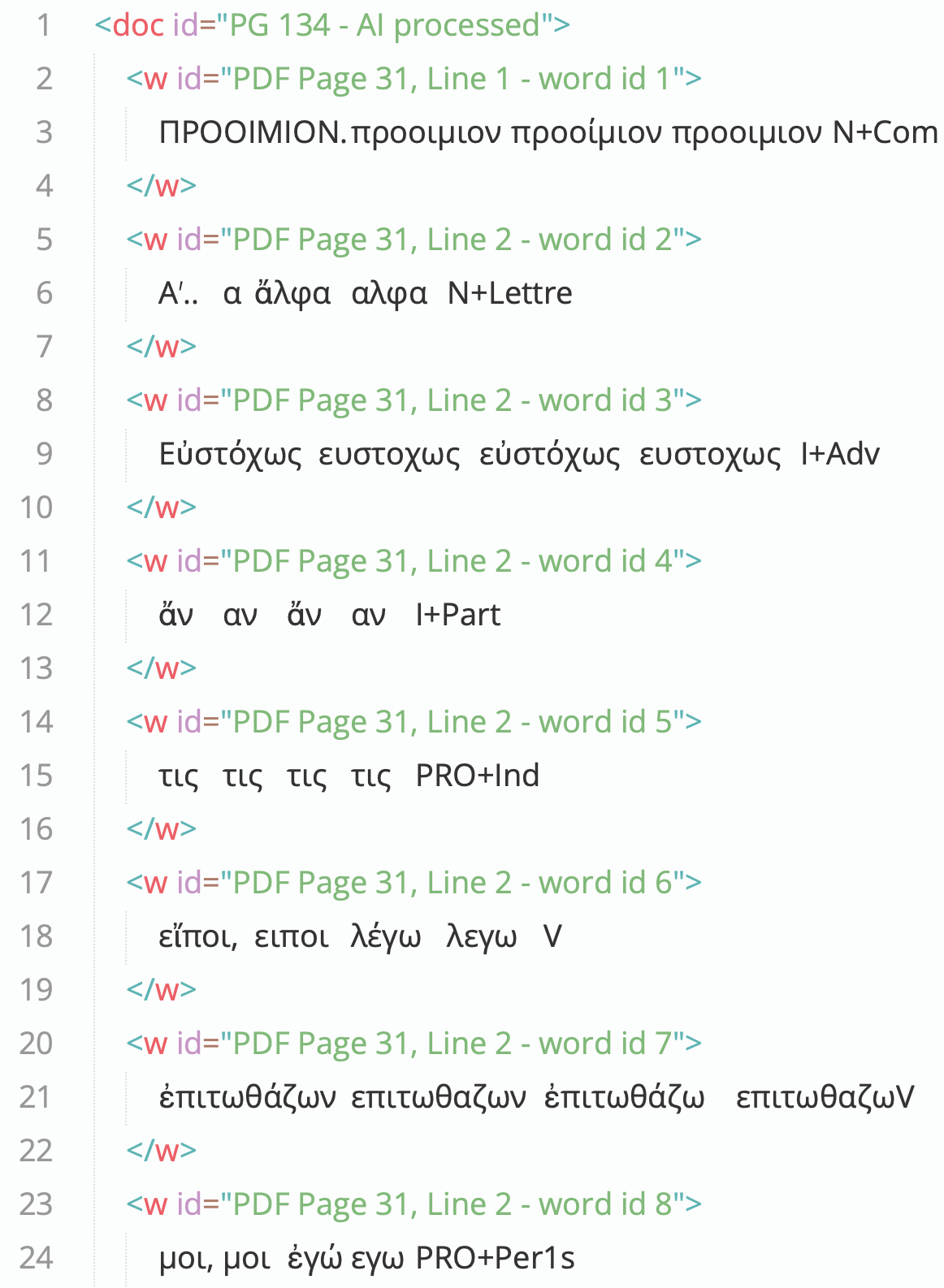}
    \caption{Example of the structured output in \texttt{.vert} format, with OCR wordform, intuitive form (clean), lemma, intuitive lemma (clean), and POS, with line IDs, word IDs and reference to PDF page.}
    \label{fig:vert-example}
\end{figure}

Prior to linguistic analysis, OCR-processed texts undergo a preparation step involving the removal of hyphens, empty lines, and Latin characters, words, or lines not belonging to the target Greek text. As it stands, however, this operation is still carried out manually, without any assurance of completion.
Following this preparation, the output is structured in the \texttt{.vert} format used by Sketch Engine, where each word token (\texttt{<w>}) is annotated with five layers of information: the OCR wordform, the lowercase diacritic-free form (intuitive form), the lemma, the lowercase diacritic-free lemma (intuitive lemma), and a morpho-syntactic tag (see Fig.~\ref{fig:vert-example}). The intuitive form and intuitive lemma allow users to submit lexical queries using plain lowercase characters, free from diacritical marks or accents—a practical affordance for researchers less familiar with polytonic Greek orthography.

The XML structure proposed here could naturally be adapted to other environments and use cases as needed. Beyond annotation, the format preserves document, page, and line identifiers, ensuring full traceability between OCR output and its original textual context—an essential feature for philological and linguistic exploration of the PG corpus. This hierarchical representation enables direct import into corpus analysis tools such as Sketch Engine or TreeTagger-compatible environments.

The resulting corpus is directly searchable on \url{gregoriproject.com}, while the raw OCR data and structured \texttt{.vert} files are publicly released on Github to ensure transparency and reproducibility. This release includes the full OCR text, line segmentation, and layout annotations, facilitating further experimentation and benchmarking on polytonic Greek material.

\section{Conclusion}

This study presented an OCR, text normalization, and text analysis workflow dedicated to the still undigitized texts of the \textit{Patrologia Graeca}, addressing the specific challenges of poorly printed nineteenth-century polytonic Greek and complex bilingual layouts. The proposed pipeline is lightweight by design, combining YOLO-based layout detection with iterative CRNN fine-tuning on a modest PG-specific ground truth, yet achieves a CER of 1.05\% and a WER of 4.69\%, improving by more than five points over existing CER baselines. In addition to OCR and layout annotations, the project releases a six-million-word lemmatized corpus, openly available for reuse (silver corpus). Its value lies not only in its scale but in its lexical and stylistic distinctiveness. These resources provide a consistent and verifiable basis for further work in historical OCR, text normalization, and the adaptation of large language models to Ancient Greek.

\section{Data Availability}

Annotations for OCR training and the corpus (raw outputs, and VERT files) are available on \url{https://github.com/calfa-co/Patrologia-Graeca}. Currently, only sample VERT files are provided due to the large size of the repository. The full set is released on Zenodo (\url{https://zenodo.org/records/15780625}). The corpus is freely accessible on \url{https://www.gregoriproject.com}.

\section{Acknowledgements}

This research was supported by: ASBL Byzantion, Calfa (Paris), UCLouvain - CIOL - Centre d'études orientales - Institut orientaliste de Louvain, UCLouvain - FSS - Fondation Sedes Sapientiae, UCLouvain GREgORI Project, UCLouvain - INCAL - Institut des Civilisations Arts et Lettres UCLouvain - RSCS - Institut de recherche pluridisciplinaire Religions Spiritualités Cultures Sociétés, and other (anonymous) private funders.
We thank Professor Jean-Marie Auwers (UCLouvain, RSCS), Professor Sébastien Moureau (UCLouvain/CIOL) and Doctor Véronique Somers (UCLouvain/CIOL) for their contributions to data curation, annotation, and evaluation.

\section{Bibliographical References}
\bibliographystyle{lrec2026-natbib}
\bibliography{lrec2026-example}

\section{Language Resource References}
\label{lr:ref}
\bibliographystylelanguageresource{lrec2026-natbib}
\bibliographylanguageresource{languageresource}

\newpage

\section*{Appendices}
\appendix

The PG comprises 161 volumes, numbered from 1 to 161 (volumes 7, 16, 86 and 87 being divided into two or three parts). All
these volumes are available in PDF format (files are listed, e.g., on the portal \href{https://patristica.net/graeca}{Patristica.net} or on the blog
\href{https://www.roger-pearse.com/weblog/patrologia-graeca-pg-pdfs}{roger-pearse.com}). At the beginning of the project, 55 volumes were selected,
based on their presence, or not, in the TLG corpus. As of now, some texts have been excluded because they were in the
meantime uploaded into the TLG corpus, or because their layout was quite specific, with more than two columns or including
languages other than Latin or Greek, such as Syriac (see, e.g., vol. 1; \url{https://books.google.be/books?id=qxANpCDQIjIC}, p. 102 and 198. The 32 volumes that were ultimately selected and are currently being processed in the project are listed in Table~\ref{tab:table_PG_files} below.

\begin{table}[ht]
\centering
\footnotesize
\resizebox{\columnwidth}{!}{\begin{tabular}{@{}rlll r@{}}
\toprule
\textbf{PG Vol.} & \textbf{URL} & \textbf{Date} & \textbf{Word Count} \\
\midrule
3    & \href{http://books.google.com/books?id=ALfUAAAAMAAJ}{\texttt{books.google.com/.../ALfUAAAAMAAJ}}            & 5th--6th AD (?) & 134{,}866 \\
5    & \href{http://books.google.com/books?id=PIQe9iqBoeQC}{\texttt{books.google.com/.../PIQe9iqBoeQC}}            & 1st--2d AD      & 46{,}164  \\
6    & \href{http://books.google.com/books?id=NZLYAAAAMAAJ}{\texttt{books.google.com/.../NZLYAAAAMAAJ}}            & 2d AD           & 170{,}482 \\
8    & \href{http://books.google.com/books?id=BwcRAAAAYAAJ}{\texttt{books.google.com/.../BwcRAAAAYAAJ}}            & 2d--3rd AD      & 168{,}277 \\
9    & \href{http://books.google.com/books?id=QAgRAAAAYAAJ}{\texttt{books.google.com/.../QAgRAAAAYAAJ}}            & 2d--3rd AD      & 82{,}135  \\
16.3 & \href{http://books.google.com/books?id=mAsRAAAAYAAJ}{\texttt{books.google.com/.../mAsRAAAAYAAJ}}            & 3rd AD          & 60{,}921  \\
21   & \href{http://books.google.com/books?id=BwcRAAAAYAAJ}{\texttt{books.google.com/.../BwcRAAAAYAAJ}}            & 3rd--4th AD     & 236{,}625 \\
42   & \href{http://books.google.com/books?id=QYHYAAAAMAAJ}{\texttt{books.google.com/.../QYHYAAAAMAAJ}}            & 4th--5th AD     & 161{,}237 \\
67   & \href{http://books.google.com/books?id=WivbHxo0L-sC}{\texttt{books.google.com/.../WivbHxo0L-sC}}           & 4th--6th AD     & 170{,}445 \\
71   & \href{https://books.google.be/books?id=worYAAAAMAAJ}{\texttt{books.google.be/.../worYAAAAMAAJ}}             & 4th--5th AD     & 210{,}957 \\
73   & \href{http://books.google.com/books?id=ywsNQz1fTewC}{\texttt{books.google.com/.../ywsNQz1fTewC}}           & 4th--5th AD     & 191{,}303 \\
87.1 & \href{http://books.google.com/books?id=CMcUAAAAQAAJ}{\texttt{books.google.com/.../CMcUAAAAQAAJ}}            & 5th--6th AD     & 151{,}167 \\
107  & \href{http://books.google.com/books?id=Ru4GAAAAQAAJ}{\texttt{books.google.com/.../Ru4GAAAAQAAJ}}            & 9th--10th AD    & 196{,}727 \\
109  & \href{http://books.google.com/books?id=Z0naYVT0w-EC}{\texttt{books.google.com/.../Z0naYVT0w-EC}}           & 9th--10th AD    & 148{,}584 \\
112  & \href{http://books.google.com/books?id=nyNKAAAAcAAJ}{\texttt{books.google.com/.../nyNKAAAAcAAJ}}            & 10th AD         & 129{,}556 \\
113  & \href{http://books.google.com/books?id=Z_QUAAAAQAAJ}{\texttt{books.google.com/.../Z\_QUAAAAQAAJ}}          & 10th AD         & 104{,}371 \\
118  & \href{http://books.google.com/books?id=j_QUAAAAQAAJ}{\texttt{books.google.com/.../j\_QUAAAAQAAJ}}          & 11th AD (?)     & 208{,}448 \\
121  & \href{http://books.google.com/books?id=PIQe9iqBoeQC}{\texttt{books.google.com/.../PIQe9iqBoeQC}}           & 11th AD         & 160{,}853 \\
122  & \href{http://www.archive.org/details/patrologiaecurs61migngoog}{\texttt{archive.org/.../patrologiaecurs61migngoog}} & 11th AD  & 150{,}647 \\
123  & \href{http://books.google.com/books?id=-SFJAAAAcAAJ}{\texttt{books.google.com/.../-SFJAAAAcAAJ}}           & 11th--12th AD   & 208{,}024 \\
124  & \href{http://books.google.com/books?id=AccUAAAAQAAJ}{\texttt{books.google.com/.../AccUAAAAQAAJ}}            & 11th--12th AD   & 210{,}302 \\
125  & \href{http://books.google.com/books?id=Z7_UAAAAMAAJ}{\texttt{books.google.com/.../Z7\_UAAAAMAAJ}}          & 11th--12th AD   & 172{,}696 \\
126  & \href{http://books.google.com/books?id=eTYRAAAAYAAJ}{\texttt{books.google.com/.../eTYRAAAAYAAJ}}            & 11th--12th AD   & 164{,}706 \\
134  & \href{http://books.google.com/books?id=DrvUAAAAMAAJ}{\texttt{books.google.com/.../DrvUAAAAMAAJ}}            & 11th--12th AD   & 196{,}859 \\
139  & \href{http://books.google.com/books?id=lscUAAAAQAAJ}{\texttt{books.google.com/.../lscUAAAAQAAJ}}            & 12th--13th AD   & 134{,}703 \\
146  & \href{http://books.google.com/books?id=xCJKAAAAcAAJ}{\texttt{books.google.com/.../xCJKAAAAcAAJ}}            & 13th--14th AD   & 156{,}848 \\
148  & \href{http://books.google.com/books?id=IlWwi1vmdb4C}{\texttt{books.google.com/.../IlWwi1vmdb4C}}           & 13th--14th AD   & 234{,}855 \\
151  & \href{http://books.google.com/books?id=PyNKAAAAcAAJ}{\texttt{books.google.com/.../PyNKAAAAcAAJ}}            & 13th--14th AD   & 399{,}518 \\
153  & \href{http://books.google.com/books?id=dIrYAAAAMAAJ}{\texttt{books.google.com/.../dIrYAAAAMAAJ}}            & 13th--14th AD   & 230{,}239 \\
155  & \href{http://books.google.com/books?id=_McUAAAAQAAJ}{\texttt{books.google.com/.../ \_McUAAAAQAAJ}}         & 14th--15th AD   & 175{,}482 \\
157  & \href{http://books.google.com/books?id=YB8RAAAAYAAJ}{\texttt{books.google.com/.../YB8RAAAAYAAJ}}            & 15th AD         & 95{,}020  \\
158  & \href{http://books.google.com/books?id=hBIT95lCqEsC}{\texttt{books.google.com/.../hBIT95lCqEsC}}           & 12th AD         & 163{,}148 \\
\midrule
     &                                                                                                              & \textbf{Total}  & \textbf{5{,}605{,}015} \\
\bottomrule
\end{tabular}
}
\caption{List of processed volumes of the PG, url of the PDF files, date of the texts, and wordcount.}\label{tab:table_PG_files}
\end{table}

\end{document}